# DELP-DAR System for License Plate Detection and Recognition

Zied Selmi[a] *, Mohamed Ben Halima[b], Umapada Pal[c] and M.Adel Alimi[b]

[a b] REGIM-Lab: Research Groups in Intelligent Machines, University of Sfax, ENIS, BP 1173, Sfax, 3038, Tunisia
[c] Computer vision and Pattern Recognition Unit Indian Statistical Institute 203 B. T. Road, olkata-700108, India

ABSTRACT

Automatic License Plate detection and Recognition (ALPR) is a quite popular and active research topic in the field of computer vision, image processing and intelligent transport systems. ALPR is used to make detection and recognition processes more robust and efficient in highly complicated environments and backgrounds. Several research investigations are still necessary due to some constraints such as: completeness of numbering systems of countries, different colors, various languages, multiple sizes and varied fonts. For this, we present in this paper an automatic framework for License Plate (LP) detection and recognition from complex scenes. Our framework is based on mask region convolutional neural networks used for LP detection, segmentation and recognition. Although some studies have focused on LP detection, LP recognition, LP segmentation or just two of them, our study uses the maskr-cnn in the three stages.

The evaluation of our framework is enhanced by four datasets for different countries and consequently with various languages. In fact, it tested on four datasets including images captured from multiple scenes under numerous conditions such as varied orientation, poor quality images, blurred images and complex environmental backgrounds. Extensive experiments show the robustness and efficiency of our suggested framework in all datasets.

## 1. Introduction

Automatic detection and recognition of License Plates (LPs) is Frequently used in several areas such as traffic management, digital security surveillance, vehicle recognition, parking management, monitoring border crossings and searching for suspicious vehicles. The process of Automatic LP detection and Recognition (ALPR) is a complex challenge due to many factors that can be divided into two subcategories: technical factors like the various LP numbering systems, types, sizes and languages; and factors related to image quality and environmental conditions such as blurred images, poor lighting conditions, deformation, orientation, different camera angles, daylight, night, raining and complex environmental backgrounds. Several methods and techniques have been used to solve the problems of detecting and recognizing LPs. Among these methods, there have been traditional approaches like edge detection, morphological operations, character-based approaches, texture-based techniques, and statistical analyses. On the other hand, researchers have used learning and classification systems. Recently, the field of deep learning and Convolution Neural Network (CNN) based methods has been used to solve the difficulties found in the detection and recognition of LPs. Despite the use of traditional and deep learning-based methods, several difficulties and challenges are still not properly solved, especially images in complex backgrounds with multi-orientations due to the viewpoint variation in camera and multi-language characters. To resolve the problems mentioned above, we propose here an efficient system for detecting and recognize LPs.

The main contributions of our work are summarized as follows:

1. We develop a system based on Mask Region CNNs (Mask-RCNNs) to correctly detect, segment and recognize LPs with multi-orientation and multi-languages and under complex environments.

2. We present for performance evaluation our newly collected challenging database containing 610 Tunisian LP images with variable orientations, weather conditions and complex backgrounds.

3. Demonstrating that our suggested system will outperform the existing approaches concerning LP detection and character recognition.

This paper is organized as follows. Section II presents the literature review for LP detection and recognition. Section III introduces the proposed system in detail. We provide the evaluation of our system, the experimental results and the discussions in section IV. Finally, section V provides the conclusions and future work.

## 2. Related work

In this section, we briefly give some ALPR work in the literature in each process (LP detection, segmentation and recognition), and then we present the work of deep learning in this area.

### 2.1. LP detection

LP detection is a primordial phase in the identification of the characteristics of a car. It is a location of an LP in an image or a video stream. In the literature, researchers have used a lot of techniques of computer vision and image processing. Yet, there have been other authors who have opted for using machine-learning and classification methods. For that, we can categorize these methods into: edge-based, color-based, texture-based and character-based methods [1]. Usually, the edge-based approach is a method of detecting contours of a rectangle since an LP has this shape with an aspect ratio. Thus, this method is normally utilized to find these rectangles [2][3].

The goal of edge-based methods is to find regions that have a higher edge density than other areas of the image that identify an LP.

However, the change in LP brightness and area is remarkable. Hence, the authors in [4,5,6] used edge detection with morphological and mathematical operations to find the rectangles that could be LPs. To properly detect an LP, Chen and Luo [7] proposed an improvement of the Prewitt arithmetic operator utilizing the horizontal and vertical projections. In [8], the authors put forward a new line density filter method to connect regions with a high-edge density and to remove parse regions from a binary edge image in every column and row.

Among the characteristics of edge-based methods is the higher detection speed. However, we do not apply them to complex images because of their sensitivity to unwanted edges. In addition, in case they are blurry, it will be difficult to find LPs. Color-based approaches identify LPs by locating their colors in the image. Shi et al. [9] and Zayed et al. [10] proposed a color model classifier. Moreover, the authors of [11] segmented the color images using a shift algorithm in the candidate regions. After that, they classified them with or without LPs. In fact, one combination of rectangular characteristics, an aspect ratio and an edge density was utilized to determine these candidate regions. For addressing such illumination variations, the writers proposed one approach of fuzzy logic in order to recognize LP colors. As a matter of fact, LP extraction using color information was able to detect slanted and distorted LPs.

However, such an approach was sensitive to several illumination alterations and would suffer from false positives, especially when other test image parts would have the same LP colors.

Texture-based approaches have attempted to detect the desired regions of their pixel intensity distribution in LPs. The authors in [12] used Support Vector Machines (SVMs) to analyze the color characteristics of LP texture. Afterwards, they utilized a Continuous Adaptive Mean Shift (CAMShift) algorithm on the results to identify the LP areas. The authors in [13] used a wavelet transform to identify few features and details that were either vertical or horizontal in the image. In [14]-[15], the authors combined Haar-like features with adaptive boosting to obtain cascade classifiers for LP extraction. Added to that, these Haar-like features can be utilized to detect objects, so they can make a classifier invariant to the LP size, color and brightness. A cascade Adaboost was used by [16] which obtained a higher accuracy rate of LP detection.

Giannoukos et al. [17] developed a concentric sliding window algorithm to locate LPs based on the local irregularity property of the plate texture in an image. Operator context scanning was used to improve the speed of detection.

Even its boundary is deformed, an LP can be detected by texture-based methods. However, these methods are computationally complex, mainly when there are many edges.

Character-based approaches will consider an LP a string of characters while detecting it through the examination of the presence of characters with an image. In addition to that, Li et al. [18] applied at the first stage Maximally Stable Extremal Region (MSER) for extracting candidate characters in images. After that, the authors constructed a conditional random field (CRF) for the representation of relationships among LP characters. LPs would be ultimately detected by the belief propagation inference on the CRF.

In [19], the authors labeled a binary object, which had exactly a similar aspect ratio, as one character and over 30 pixels. In fact, a Hough transform was applied to detect straight lines, in a similar way for the low part of these connected objects. If both straight lines were parallel in a specific range while having the same number and characters, the authors would consider the area between them an LP. Hontani et al. [20] applied a scale-space analysis for the extraction the characters. In [21], the authors used the width of characters and the difference between character regions and backgrounds, to recognize the character region. After that, the LP was extracted by finding the inter-character distance in the plate region.

Yet, character-based methods can be more reliable. As a result, we can obtain a high recall. On the other hand, the performance may be largely affected by text in the image background.

Hybrid methods have been proposed. They have represented a combination of two or more approaches. In [22], the authors combined the texture and color characteristics by applying fuzzy rules to extract the same texture and color. Added to that, Lim et al. [23] proposed combining the sift-based unigram classifier with the MSER method.

*2.2. LP segmentation and recognition*

The character segmentation stage is the second one in ALPR systems. A lot of methods and techniques have been applied to detect the region character in an LP.

In order to detect an LP candidate region, the writers of [24] utilized the image segmentation technique, named the sliding window. In [25] the authors segmented the LP with a fuzzy logic approach and a CNN to extract the features of an LP. The authors in [26] used a connected-component-based approach for character segmentation [26]. The character contours were used to detect characters into LPs [27]. In [28] Extremal Regions (ER) were employed to segment characters from coarsely detected LPs and to refine plate location. The authors in [29] adopted MSER for character segmentation.

The last step for the ALPR systems is the LP recognition. In the Literature, there have been various recognition methods of characters, like extracted features, neural networks and raw data.

In [30] the authors put forward template matching to extract similar characters. It is important to mention similarity measurement techniques, such as Mahalanobis, Hamming and Hausdorff distances. In [31], the authors used horizontal and vertical projections to generate a feature vector. After extracting features, we can use a lot of classifiers to recognize characters, such as SVMs [32], hidden Markov models [33] and artificial neural networks [34]. Several researchers have integrated two types of classification schemes: multi-stage classification [33], and parallel mixture of different classifiers [36].

*2.3. ALPR methods based deep learning*

In recent years, the deep learning approach has been widely used in several fields. Among these domains, one finds in the literature intelligent transport systems, specifically in ALPR.

A lot of researchers have proposed a CNN-based method for LP detection and recognition. Wang et al. [37], in order to recognize characters, utilized the CNN model to training data. As a matter of fact, to detect the text, Wang et al. used CNN prediction for text/non-text. The authors in [38] developed a multi-layer CNN to spot texts in images. The authors in [39] utilized one cascade CNN to detect frontal views and LPs of vehicles. In [40], the writers developed a CNN-based method exclusively for LP detection in multi-orientations and under various conditions. An RCNN was applied with an SVM to detect LPs. Li and Chen [42] proposed CNN-based characters. The latter would be cropped from general text to detect characters.

To recognize characters, the authors used a Recurrent Neural Network (RNN) with Connectionist Temporal Classification (CTC) to label sequential data, so as to recognize whole LPs without character-level segmentation. A classifier of sparse network of Winnows was used by the authors in [43], to extract candidate regions from an LP. After that, a CNN model was applied to filter these regions. Lele et al. [44] proposed a you-only-look-once-based framework for multidirectional car LP detection. This suggested method could elegantly manage rotational problems in real-time scenarios.

Zied et al. [45] put forward one system with preprocessing steps. For example, we mention edge detection, geometric filtering etc. After that, the authors implemented a CNN model for LP / non-LP classification to detect LPs. Furthermore, digits and characters were classified and recognized from LPs by the utilization of a second CNN.

The algorithm in [46] attempted to solve both LP detection and recognition problems by a single deep neural network, which would extract features for the input image, generate the region proposal and apply the Non-Maximum Suppression (NMS) method for LP detection. For the recognition phase, Li et al. used bidirectional RNNs with CTC to label sequential features. The author in [47] proposed a method containing three phases: a preprocessor, a proposal extractor, and an LP classifier for LP detection.

**3. DELP-DAR system**

In this section, we propose an overview of our DEep-learning License Plate Detection And Recognition (DELP-DAR) system for LP detection and recognition (See Fig.1).

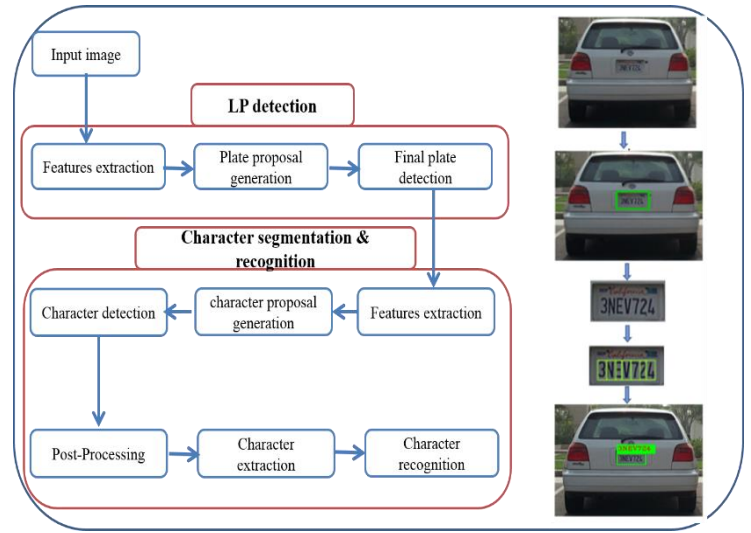

**Fig. 1.** Flowchart of DELP-DAR system

DELP-DAR is a framework that finds and reads LPs in an image. We especially focus on finding and reading LPs in unconstrained and tough situations, as approaches in this area are

still very limited. Such a framework can be split into three main parts, which are:

1. Detecting LPs
2. Segmenting and detecting each character
3. LP characters recognition

Since our system is based on Mask-RCNNs [48], we give some insights into this new technique.

*3.1. Mask-RCNN*

We briefly explain in this sub-section the core component of Mask-RCNNs as a key to our approach. Indeed, Mask-RCNNs [49] are based on faster RCNNs. CNNs have been used for object detection for years now. The very first approach is called RCNNs, which generates regions to check for objects. At first, the regions that can contain objects are determined. This is done by combining edges and other features to differentiate between objects. This approach does not involve any kind of learning and is solely based on handcrafted features. After generating regions, a CNN classifies the regions as "nothing" or "object-X". This approach works well. However, there is much room for improvement. Especially the handcrafted region proposal is not accurate enough and produces up to a few thousand of regions, which all must be classified. This makes the whole approach very slow and this is the reason for emerging approaches like fast RCNN [50] and faster RCNN [49]. The latter improves the accuracy of the whole framework and speeds up everything a lot.

The core idea is not to generate a region based on handcrafted features but to let CNNs learn to find Regions of Interest (ROIs) on its own. This is done by adding a ROI-pooling layer. This layer extracts region proposals after the convolution layers are applied to an image. Therefore, when examining a region, we benefit from two things:

- the CNN already generates good features;
- the whole image already goes through a CNN. Thus, we do not have to pass any region again through the convolution layers;

Additionally, a CNN is trained to predict a good bounding box based on the initially proposed region of the ROI pooling layer. This leads to enormous success of the faster RCNN. The Mask-RCNN, originally developed to predict segmentation on a pixel-level, improves the ROI pooling layer by improving its accuracy, hence ameliorating the accuracy of the bounding boxes.

The SoftMax layer classifies the extracted region, and the boundary box regressor refines the bounding boxes. These latter are refined based on the initial extracted window.

*3.2. License plate detection*

Figure 2 describes the flowchart of our proposed approach of LP detection. Our system is based on Mask-RCNN architecture [48], which classifies the regions extracted into "LP found" and "non-LP found". We merge several datasets for positive and negative sample generation for training.

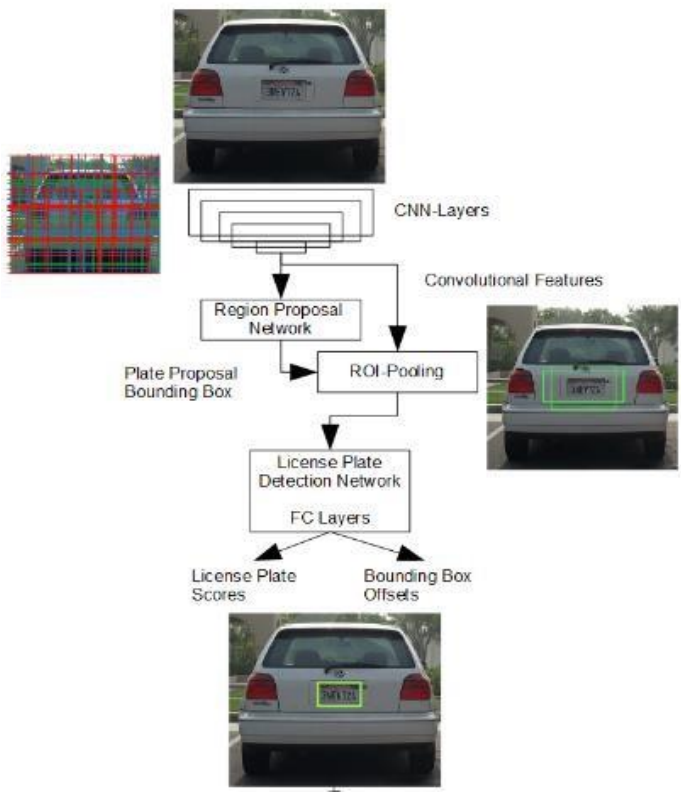

**Fig. 2.** Flowchart of LP detection

*3.2.1. Features extraction*

We use a network comparable to GoogLeNet, but we cut down the inception module [51], amount to six, and add several pooling layers by doing convolution with a stride size of two. The resulting feature maps are one-eighth size of the original input image. This enables us to have fine-grained detection of LPs even for small plates while keeping the computational load low. We use the feature maps as a base for both detection and recognition.

We utilize the novel activation function called 'swish' which is proposed by Prajit, et al. [52] from Google Brain, it's a relatively simple function that is multiplying the x input with the sigmoid function for x (see Fig.3 for swish-activation function).

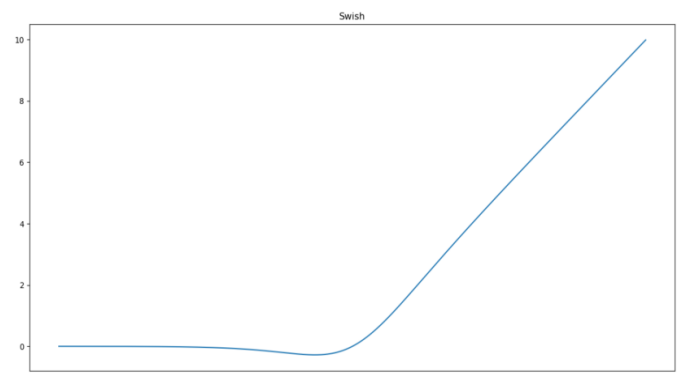

**Fig. 3.** Swish activation function

The mathematical definition of 'swish' is presented in equation 1:

$$f(x) = x * sigmoid(x) \qquad (1)$$

Swish is a smooth function. This means that it does not change direction suddenly as does RELU near x = 0. On the contrary, it bends smoothly bends from 0 towards values < 0 and then upwards again.

This observation means that it is also non-monotonic. So, it does not stay stable or move in one direction, like RELU. According to the experiments done by the authors in [52] show that Swish tends to work better than RELU on deeper models in several complex datasets. It works better than RELU with the same level of computational efficiency. In experiments with ImageNet with identical models running RELU and swish, the new function has achieved a higher classification accuracy of 0.6 to 0.9%.

The configuration of our model network for LP detection is shown in Table 1.

**Table 1. Configuration of LP detection network**

| No | Layer | Filters | Size | Input | Output |
|---|---|---|---|---|---|
| 1 | conv-swish | 64 | 3x3 | 960x570 | 960x570 |
| 2 | max-pooling | 64 | 3x3 | 960x570 | 481x286 |
| 3 | conv-swish | 64 | 1x1 / 1 | 481x286 | 481x286 |
| 4 | conv-swish | 64 | 3x3 / 1 | 481x286 | 481x286 |
| 5 | conv-swish | 64 | 3x3 / 2 | 481x286 | 481x286 |
| 6 | max-pooling | 64 | 3x3 | 481x286 | 242x144 |
| 7 | Incept-swish | 32 | 3x3 | 242x144 | 242x144 |
| 8 | max-pooling | 64 | 3x3 | 242x144 | 122x73 |
| 9 | Incept-swish | 48 | 3x3 | 122x73 | 122x73 |
| 10 | conv-swish | 128 | 1x1 / 1 | 122x73 | 122x73 |
| 11 | conv-swish | 128 | 3x3 / 1 | 122x73 | 122x73 |
| 12 | conv-swish | 128 | 3x3 / 2 | 122x73 | 122x73 |
| 13 | Incept-swish | 64 | 3x3 | 122x73 | 122x73 |
| 14 | Incept-swish | 96 | 3x3 | 122x73 | 122x73 |
| 15 | conv-swish | 96 | 1x1 / 1 | 122x73 | 122x73 |
| 16 | conv-swish | 96 | 3x3 / 1 | 122x73 | 122x73 |
| 17 | conv-rpn | 512 | 3x3 | 122x73 | 122x73 |
| 18 | fc-swish-roi | - | - | 8x7 | 2048 |
| 19 | fc-swish-roi | - | - | 2048 | 2048 |
| 20 | fc-swish-bb | - | - | 2048 | 8 |
| 21 | fc-swish-cls | - | - | 2048 | 2 |

### 3.2.2. *Plate proposal generation*

For plate proposal generation, which can generate candidate objects in images, we use a Mask-RCNN without the segmentation stage, but solely the Roi-ALign layer. A Regional Proposal Network (RPN) is a convolutional network and a fully connected classifier, which takes convolutional features as an input and outputs a set of bounding boxes. It can be trained end-to-end. In this work, we modify the Mask-RCNN slightly to make it suitable for car LP detection. According to the scales and aspect ratios of LPs in the wild, we introduce k=12 anchor boxes, which are applied at each position of the input feature maps. The anchor boxes are as follows:

**Table 2. Anchor boxes**

| Width | Height |
|---|---|
| 64 | 64 |
| 48 | 128 |
| 64 | 192 |
| 96 | 256 |
| 128 | 352 |
| 128 | 128 |
| 192 | 512 |
| 256 | 704 |
| 352 | 832 |
| 256 | 256 |
| 512 | 512 |
| 192 | 832 |

The features are concatenated along the channel axis and form a 512-d feature vector, which is then fed into two separate fully convolutional layers for LP/non-LP classification and bounding box regression. In order to avoid the duplicate RoIs and RoIs having a lower likelihood score, NMS is implemented in case the Intersection over Union (IoU) overlap is 0.7.

### 3.2.3. *Final plate detection*

The last step of our system is to filter the proposed regions into non-LP and LP regions. Therefore, this step is similar to the one before, but now it tries to minimize false positives and negatives.

For this, we use the RoI-Align layer proposed by the Mask-RCNN and set the pooling size to 8x7. We increase "fine-grained" in the X direction, because LPs tend to be much wider rather than higher.

Two fully connected layers with 2,048 neurons and a dropout rate of 0.1 are used to extract discriminative features for LP detection. The features from each RoI are flattened into a vector and passed through fully connected layers. The plate classification layer has two outputs, which indicate the SoftMax probability of each RoI as LP/non-LP. The plate regression layer outputs the bounding box coordinate offsets relative to the anchor box for each proposal, as in standard faster RCNNs.

### 3.3. LP segmentation and recognition

Once the LP is detected, we will begin the steps of character recognition. However, before this latter, we need to segment and isolate each character from an LP, and finally combine the results (see Fig 4). Here are the steps to follow:

1. Extract possible character areas and classify them by Mask-RCNN.

2. Remove areas which are too wide/small to be a character.

3. Cluster areas which all have nearly the same height and y coordinate.

4. Recognize the characters with Mask-RCNN boxes.:

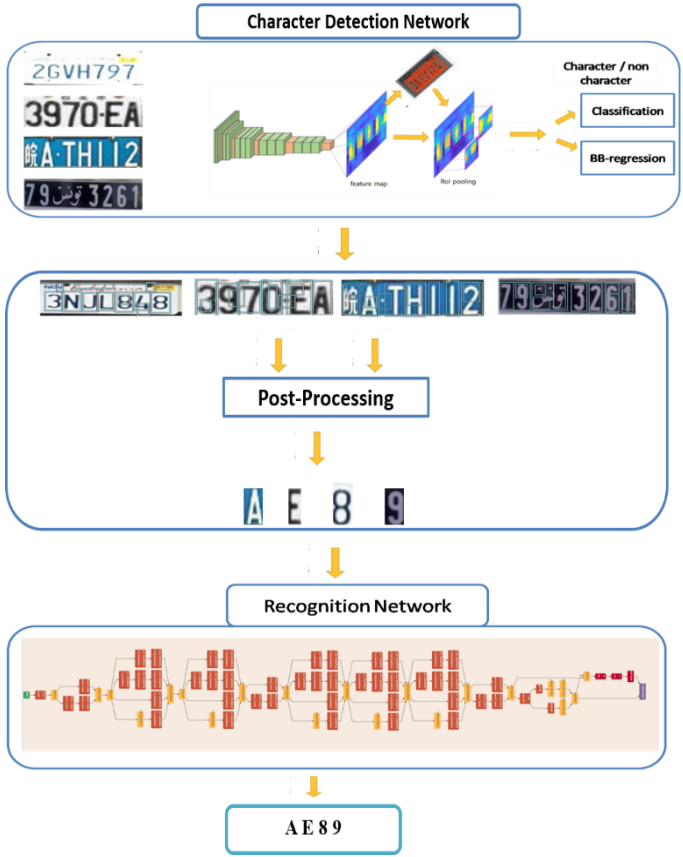

**Fig. 4.** Architecture of LP segmentation and recognition

Character segmentation is a significant phase for facilitating the recognition process. Indeed, this comprises the extraction of characters from the LP image. Many factors make this stage complex. For instance, we state the noise, the rotate LP, the low blur resolution, the different colors, and the numbering system.

To deal with these complexities, we develop one novel process so as to effectively segment the characters.

Our segmentation system is divided into two parts. The first one is to detect each character in an LP using a Mask-RCNN-based method to classify characters and non-characters. The used method gives a prediction result for characters and non-characters.

This prediction is based on a training set that contains correct and incorrect annotation characters from the datasets and our generated artificial dataset.

The second part is designed to improve character detection in each LP. After applying the Mask-RCNN, we remove regions, which are too wide or too small to be a character. We remove every region, where the width of a character is bigger than the height of the same character or if the characters are smaller than the 1/12 of the image size.

The last step of area extraction is to extract a set of regions, which vary in size and in their y coordinate only by small amount d where d is the distance between the y coordinates and the heights of two succeeding characters and here is the formula (see equation 2):

$$d = |y1 - y2| \text{ et } d = |h1 - h2| \qquad (2)$$

If $\frac{d}{imageheight} > d\_max$ Where d_max is a maximum distance between two characters, and to ensure that we will extract a character and after several tests have shown that the most stable and the best value on several scenarios and images is 0.1 because the more we d_max fixed at a high level, more the variance of the height and y coordinates of two characters or boundary box can be varied until they are ignored. Then the regions are not grouped into the current cluster.

We set every region as a starting point for clustering and we redo the whole cluster-pipeline. After doing this, for every start-region, we choose the one cluster with the corresponding start region, which has at least four but less than 10 regions and has the lowest deviation of height and y coordinate.

To obtain a good result in character recognition, we need a lot of characters. For that, we collect 38-character classes (0-9, A-Z+ one Arabic word + negative class) for the training of our model.

We exploit two other datasets to extract character and non-character images, which are ICDAR 2003 [58] and Chars74 [59]. Furthermore, we crop the characters ourselves from the LP Caltech, PKU, AOLP and Tunisian datasets. Moreover, we create 3k artificial LPs to the order of 90K characters in an uppercase format and digits, and 50k non-characters.

To raise the number of each class, we use the technique of data augmentation, namely: adding random Gaussian noise, changing random brightness, swapping color channels, applying a median filter, and randomly rotating characters in multiple angles (+10, -10, +15,-15,+20,-20), in order to obtain 310k characters and 160k non characters.

The model used in the recognition process is the same as that of LP detection whose modification is in the input image. Table 3 describes the configuration of our model network.

### 3.4. Loss functions and training

The faster R-CNN was previously trained using a multi-step approach, training parts independently and merging the trained weights before a final full training approach. Since then, it has been found that doing end-to-end joint training leads to better results.

After putting the complete model together, we end up with four different losses, two for the RPN (background, foreground classification and bounding box regression) and two for the R-CNN (final class output and final bounding box regression). This yields an overall loss of :

$$L_{overall} = L_{cls\,RPN} + L_{bb\,RPN} + L_{cls\,RCNN} + L_{bb\,RCNN} \qquad (3)$$

where $L_{cls\,RPN}$ is the region proposal network loss for distinguishing between the background and the foreground, $L_{bb\,RPN}$ is the RPN loss for refining bounding boxes, $L_{cls\,RCNN}$ is the RCNN loss for distinguishing between the background and the different classes, and $L_{bb\,RCNN}$ is the RCNN loss for refining bounding boxes.

The class losses are mainly cross entropy soft max losses and L2 for bounding box regression, as follows:

$$L_{cls}(y, p) = \sum_{i=0}^{N} y_i \log(p_i) \qquad (4)$$

where N is the amount of different classes, y is the labels, $y_i$ is 1 if and only if the sample is of class i and pi is of class i.

The loss for the bounding box regression can be a simple L2 loss as follows :

$$Lbb = \sum_{i=0}^{N}(x_{true} - x_{predicate})^2 \quad (5)$$

**Table 3. Configuration of character recognition**

| N° | Layer | Filters | Size | Input | Output |
|---|---|---|---|---|---|
| 1 | conv-swish | 64 | 3x3 | 530x300 | 532x302 |
| 2 | conv-swish | 64 | 1x1 / 1 | 532x302 | 532x302 |
| 3 | conv-swish | 64 | 3x3 /1 | 532x302 | 532x302 |
| 4 | conv-swish | 64 | 3x3 /2 | 532x302 | 532x302 |
| 5 | max-pooling | 64 | 3x3 | 530x300 | 267x152 |
| 6 | incept-swish | 32 | 3x3 | 267x152 | 267x152 |
| 7 | conv-swish | 128 | 1x1 / 1 | 267x152 | 267x152 |
| 8 | conv-swish | 128 | 3x3 /1 | 267x152 | 267x152 |
| 9 | conv-swish | 128 | 3x3 /2 | 267x152 | 267x152 |
| 10 | max-pooling | 64 | 3x3 | 135x77 | 135x77 |
| 11 | incept-swish | 64 | 3x3 | 135x77 | 135x77 |
| 12 | conv-swish | 96 | 1x1 / 1 | 135x77 | 133x75 |
| 13 | conv-swish | 96 | 3x3 /1 | 135x77 | 133x75 |
| 14 | con-rpn | 512 | 3x3 | 133x75 | 133x75 |
| 15 | fc-swish-roi | - | - | 7x7 | 2048 |
| 16 | fc-swish-roi | - | - | 2048 | 2048 |
| 17 | fc-swish-bb | - | - | 2048 | 8 |
| 18 | fc-swish-cls | - | - | 2048 | 38 |

## 4. Experimentation results and evaluation

In this section, we perform experiments to validate and verify the robustness and efficiency of our ALPR framework. The experiments are conducted on an Intel PC Core i7 CPU2 GHz, 8 GB of RAM and a ubuntu LTS 16 operating system. Our proposed system is implemented with python, opencv 3.1 and a caffe framework, which are performed on NVIDIA GeForce GPU with 8 Go memory.

### 4.1. Datasets

To evaluate the performance and robustness of our proposed DELP-DAR framework, we use four datasets.

The first dataset is the Application-Oriented LP (AOLP) benchmark [30]. This latter contains exactly 2,049 images of Taiwan LPs, which are divided to three categories: Road Patrol (RP) with 611 images, Law Enforcement (LE) with 757 images and Access Control (AC) with 681 images. The images are captured from various locations and times, in different weather and illumination conditions.

The second dataset is the PKU (Peking University) benchmark [8] which contains 3,977 images with Chinese LPs captured under diverse scenarios. It is categorized into five groups (G1-G5) corresponding to different configurations, environments, times, etc. The resolution images in G1, G2 and G3 is 1028x728 pixels. In G4 and G5, the resolution is respectively 1600x1232 and 1600x1200 pixels. We also notice that in the G1-G4 sub-category, there is only one car in the image and consequently only one LP. On the other hand, sub-category G5 contains multiple LPs in a single image.

The third dataset is Caltech benchmark [53]. This dataset contains126 images with Americain LPs with a 896x592 resolution. These images are taken in Caltech parking's in daytime and under complex background. In this dataset, only one LP exists in a single image.

To validate our system with multiple language LPs, we conduct tests and experiments on the latest dataset, which represent Tunisian LPs containing Arabic characters. The Tunisian dataset Has 740 images collected by Kteta et al.[54]. Considering the limitations in the number and complexity images of this dataset, we collect by ourselves an extension dataset for Tunisian LPs.

This new extension dataset contains 610 images, taken by a phone with a multiple resolution (320x240, 640 x480 and 1280x700 pixels). These images are taken in various camera viewpoints, in different illumination conditions and with complex backgrounds (see Fig. 5).

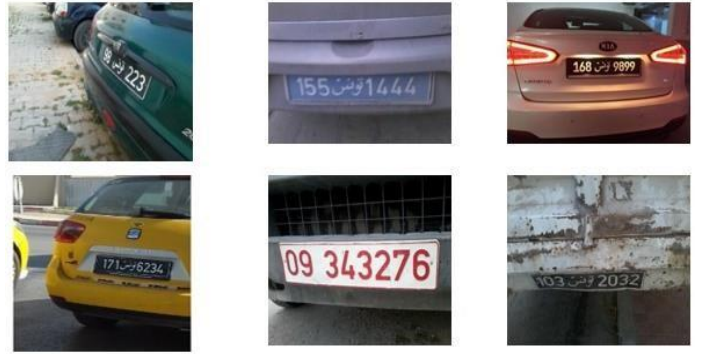

**Fig. 5.** Examples of new Tunisian dataset

In the new Tunisian dataset, there are several LP types, colors (black: 60%, blue: 20%, white; 20%), fonts and sizes.

To increase the size of training image, we use the technique of data augmentation for each sub-category of all datasets by utilizing transformation, rotation, Gaussian blurring and random lighting, to obtain 265 k true LPs and 140 k non-LPs cropped from several datasets with the application of data augmentation.

### 4.2. Evaluation Criterion

In this section, we describe the different metrics used to evaluate the effectiveness of our method in LP detection and recognition with the existing state-of-the-art methods. In the first protocol evaluation (LP detection), we follow the criterion used in [8], which uses precision and recall metrics as follows:

$$\text{Precision} = TP/ (TP + FP) \quad (6)$$

$$\text{Recall} = TP / (TP + FN) \quad (7)$$

where TP represents detected true positive boundary box LPs (correct LPs), FP represents false positive boundary box LPs detected (incorrect LPs), and FN represents false negative LPs failed to be detected.

Another evaluation criterion for the accuracy rate of LP detection is used in our approach, inspired by yuang [8], which sets the detection ratio. We assume that each LP is completely encompassed by the bounding box and that PR∩GT/PR∪GT ≥ 0.5, where PR is a predicate rectangle and GT is a ground-truth LP.

In a Tunisian older dataset, the authors in [54] used the accuracy rate for LP detection, which consisted of true positive and true negative divided by true positive, true negative, false positive and false negative boundary boxes.

For the recognition evaluation protocol, we evaluate it by character recognition accuracy defined as the number of correctly recognized characters divided by the total number of ground truth.

### 4.3. Performance evaluation in LP detection

In this section, we compare the LP detection performance of our DELP-DAR system with other state-of-the-art methods.

In Fig 6, we illustrate the process of LP detection that shows the different steps of LP detection, where the LP detection results are presented on each dataset separately: in Table 4 for the AOLP dataset [30], in Table 5 for the PKU dataset [8], and in Table 6 for Caltech dataset [53]. Finally, the evaluation for the older (TLPs 1) [54] and the newer Tunisian dataset (TLPs 2) is provided in Table 7.

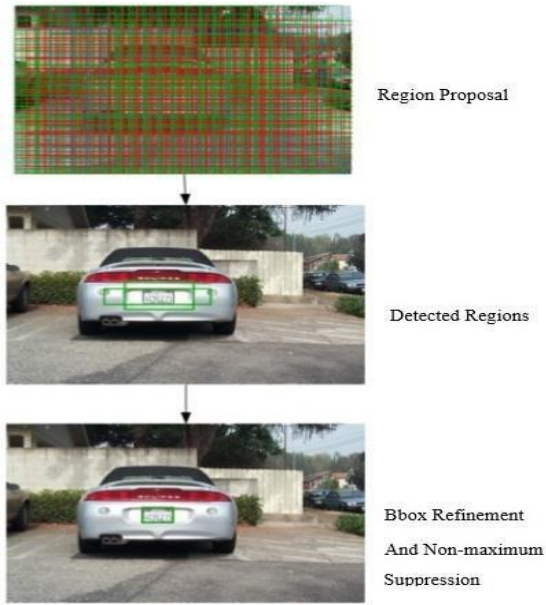

**Fig. 6.** Steps of LP detection

For the first AOLP dataset, we notice that our DELP-DAR system, which is based on the Mask-RCNN, is more performing at the precision level compared to the state-of-the-art method that achieved 99.3 % in AC, 99.2 % in LE and 98.9 % in RP (see Table 4). Hsu et al [8] used the edge detection method and the EM algorithm for clustering and obtained 91% results in AC and 91% in both LE and RP. Zied et al [45] obtained 92.6% in AC, 93.5 % in LE and 92.9 % in RP. The authors in [45] used pre-processing steps like the geometric filter and edge detection before moving to classifying rectangles to decide whether it was a correct LP.

Li [42] used CNN-based text string detection and classification for true and false bounding boxing. This method obtained 98.5 % as a precision result in AC, 97.7 % in LE and 95.2 % in RP.

For the recall result, we remark that our model is better compared to the first three authors: Hsu et al., Zied et al. and Li et al. with: 99.4%, 99.2 % and 98.9 % respectively, in the three AC, LE and RP. On the other hand, our framework is not better than Li et al. [46] system that used CNN feature extraction.

The authors obtained results as follows in AC, LE and RP: 99.5 %, 99.3 % and 98.8% respectively. As we notice, the values of our results and those of the authors in [46] are very convergent.

In Fig.7, there are different examples of LP detection in the AOLP dataset for each subcategory.

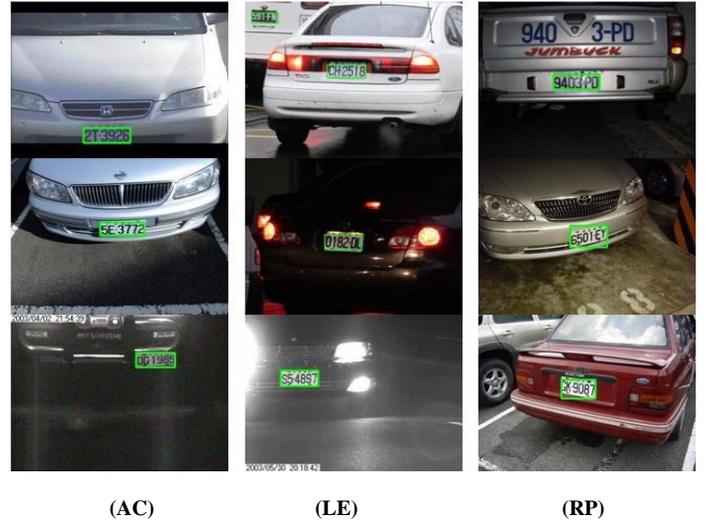

      **(AC)**      **(LE)**      **(RP)**

**Fig. 7.** Example for LP detection result in AOLP dataset by our DELP-DAR system.

**Table 4. Table: Experimental results on AOLP dataset for LP detection. (P=Precision, R=Recall)**

| Detection performance (%) | | | | | | |
|---|---|---|---|---|---|---|
| Subset<br>Methods | AC | | LE | | RP | |
| | P | R | P | R | P | R |
| Hsu *et al.* [30] | 91.0 | 96.0 | 91.0 | 95.0 | 91.0 | 94.0 |
| Zied *et al.* [45] | 92.6 | 96.8 | 93.5 | 93.3 | 92.9 | 96.2 |
| Li *et al.* [42] | 98.5 | 98.3 | 97.7 | 97.6 | 95.2 | 95.5 |
| Li *et al.* [46] | - | 99.5 | - | 99.3 | - | 98.8 |
| DELP-DAR (ours) | 99.3 | 99.4 | 99.2 | 99.2 | 98.9 | 98.8 |

The detection results for the PKU dataset compared with other methods [55], [56], [8] and [46] are presented in Table 5. As seen from this table, our proposed method achieves a higher average detection ratio of 99.4 % compared with the result in [55],[56], [8]. On the other hand, the authors in [46] achieved 99.8 %, which is slightly a higher average detection ratio as regards all the subcategories of the PKU dataset.

The images in Fig.8 show some examples with LP detection by our system in PKU dataset.

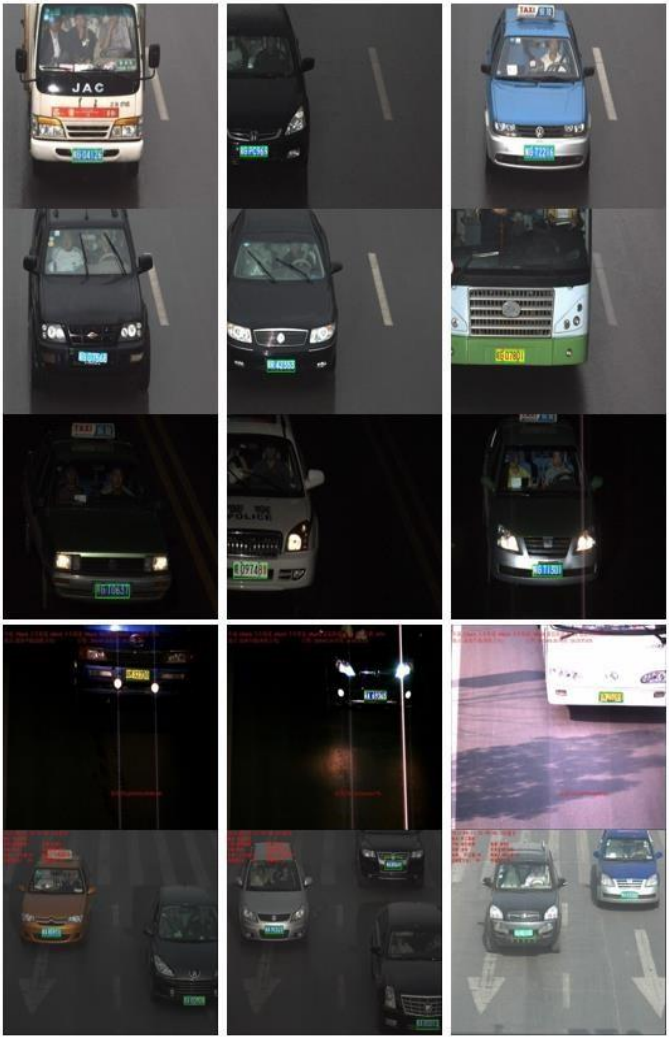

**Fig. 9.** Example for LP detection result in PKU dataset by our DELP- DAR system images organized in five subcategories (G1-G5) respectively from line 1 to line 5

**Table 5. Experimental results on PKU dataset for LP detection**

| | Detection performance (%) | | | | | |
|---|---|---|---|---|---|---|
| Subset<br>Methods | G1 | G2 | G3 | G4 | G5 | Average |
| Zhou *et al.* [55] | 95.4 | 97.8 | 94.2 | 81.2 | 82.0 | 90.2 |
| Li *et al.* [56] | 98.8 | 98.4 | 95.8 | 81.1 | 83.3 | 91.5 |
| Yuan *et al.* [8] | 98.7 | 98.4 | 97.7 | 96.2 | 97.3 | 97.6 |
| Li *et al.* [46] | 99.8 | 99.8 | 99.8 | 100 | 99.3 | 99.8 |
| DELP-DAR (ours) | 99.5 | 99.4 | 99.4 | 99.6 | 99.1 | 99.4 |

As given in Table 6, which describes a comparison with four other methods for the Caltech dataset, we remark that our approach achieves higher precision and recall of 98.9 % and 98.6 % respectively. The images of Fig. 9 depict some detection results by our system in Caltech dataset.

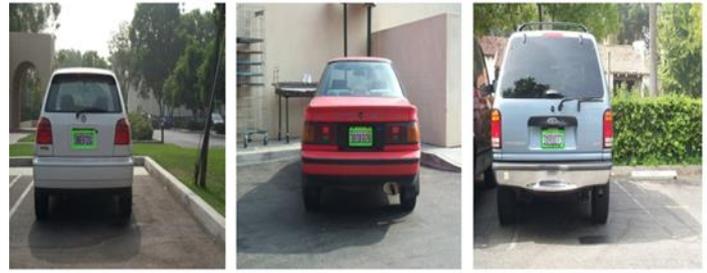

**Fig. 9.** Example for LP detection result in Caltech dataset

**Table 6. Experimental results on CALTECH dataset for LP detection. (P=Precision, R=Recall)**

| | Detection performance (%) | |
|---|---|---|
| Methods | P | R |
| Lim & Tay [57] | 83.7 | 90.4 |
| Zhou *et al.* [55] | 95.5 | 84.8 |
| Zied *et al.* [45] | 93.8 | 91.3 |
| Li *et al.* [42] | 97.5 | 95.2 |
| DELP-DAR (ours) | 98.9 | 98.6 |

The results of our framework in Tunisian datasets, as given in Table 7, describe a comparison with the Kteta methods [54], who used edge detection, projection and morphological filters in the detection phase. In the recognition phase, the author used an averaged pixel method to extract features to generate a vector for each character. Our DLLP-DAR is the best in the accuracy rate with 97.5 % compared to older datasets (TLPs1). In new collected Tunisian datasets (TLPS 2), we achieve a higher accuracy rate of 97.9 %.

The images of Fig.10 represent some examples in LP detection results by our DELP-DAR framework in Tunisian datasets.

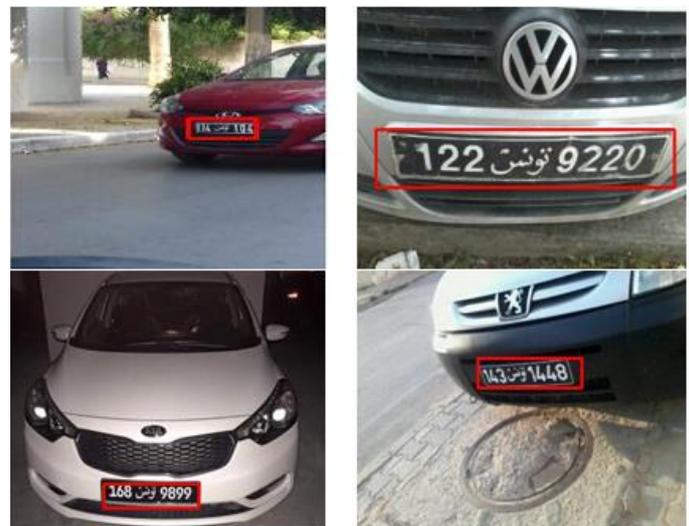

**Fig. 10.** Example for LP detection result in Tunisian dataset

**Table 7. Experimental results on TUNISIAN dataset for LP detection**

*4.4. Analysis & discussion for LP detection*

| Detection performance (%) | | |
|---|---|---|
| Subset<br>Methods | Accuracy rate | |
| | TLPs1 | TLPs2 |
| **Kteta *et al.* [54]** | 88.6 | - |
| **DELP-DAR(ours)** | 97.5 | 97.9 |

We notice in the phase of LP detection that our approach obtains better results in all the datasets in the criterion of precision evaluation for all state-of-the-art authors. These precision results confirm the robustness and efficiency of our framework.

At the recall level, our approach also gets better results in Caltech, as well as very high values in the accuracy rate for the Tunisian dataset.

However, in the other two datasets, we get better results compared with the three authors in the AOLP dataset and four authors in the PKU dataset. The small failure of our results appears with the system of [46] whose approach of LP detection consisted in feature map extraction through CNNs, generation of RPNs and regression bounding boxes. In this phase, we notice that Li et al. approach and ours have the same development principle, which explains the reconciliation of the results in AOLP and PKU, and the difference between the two results is slight (the average difference is 0.1% in AOLP and 0.4% in PKU dataset).

In Fig.11 we show some failing images and that our system does not detect LPs. This flaw in some images is caused by the very low or very high brightness in the image (image 1 in column 1, and image 2 column 2). In column 3, we remark that the image contains three vehicles, hence three LPs, but our system detects only two LPs in some cases.

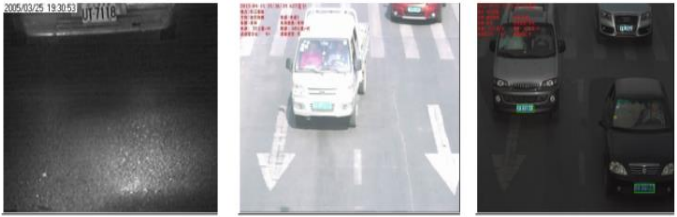

**Fig. 11.** Failure example of LP detection

4.5. Performance evaluation in LP segmentation and recognition

In the segmentation part, we label each character into an LP from all the training datasets, as well as from the generated artificial LP. The result obtained from the character detection in this phase is very high in Caltech, PKU, AOLP and Tunisian datasets as follows: 97.7 %, 98. 8 %, 97.8 % and 97.4 %, respectively.

The character recognition accuracy rate of our method shows higher accuracy compared with [42] and [45] in the Caltech dataset where our result achieves 97.2%, as presented in Table 8. The AOLP dataset with all the sub-categories (AC, LE and RP), our method obtains again better results compared to the work of [60],[61],[30],[42], [45] and [46], where our results respectively achieves 97.8 %, 97.4% and 96.3 % in AC, LP and RP, respectively.

The PKU dataset is intended for LP detection only. For that, the dataset created by [8] only labeled on LP's ground truth and did not provide the ground truth of the PKU dataset for character recognition. Accordingly, we have not found in the state-of-the-art researchers who evaluated character recognition. For this, and among our challenges, we realize this ground truth by ourselves to evaluate the recognition results on this dataset. Therefore, our system obtains the following good results; 98.4 %, 98 %, 97.8 %, 97.3 % and 96.9 %in G1, G2, G3, G4 and G5, respectively (see Table 9). The Chinese characters in this dataset are considered non-characters in the training phase, hence not being recognized.

For our Tunisian dataset, LPs are a bit specific because these contain Arabic characters such as (تونس) in the middle of the plate.

To recognize this Arabic word, we extract this word in an image form from our dataset for training, because the characters in Arabic are bound together and not separate characters.

As presented in Table 10, our model is also the best in the old dataset collected by Kteta et al. [54], which gives us a good result of 96.9%. Besides, the results in our new dataset are good: 97.5%.

**Table 8. Comparison with other character recognition methods on subsets of AOLP and Caltech dataset**

| Recognition performance (%) | | | | |
|---|---|---|---|---|
| Subsets<br>Methods | Caltech | AC | LE | RP |
| **Jia *et al.* [60]** | - | 90 | 86 | 90 |
| **Christos *et al.* [61]** | - | 92 | 86 | 91 |
| **Hsu *et al.* [30]** | - | 95 | 93 | 94 |
| **Li *et al.* [42]** | 92 | 94.8 | 94.1 | 88. |
| **Zied *et al.* [45]** | 94.8 | 96.2 | 95.4 | 95.1 |
| **Li *et al.* [46]** | - | 95.29 | 96.57 | 83.6 |
| **DELP-DAR (ours)** | 97.2 | 97.8 | 97.4 | 96.3 |

**Table 9. Comparison with other character recognition methods on**

| Recognition performance (%) | | | | | |
|---|---|---|---|---|---|
| Subsets<br>Method | G1 | G2 | G3 | G4 | G5 |
| **DELP-DAR (ours)** | 98.4 | 98.0 | 97.8 | 97.3 | 96.9 |

subsets of PKU dataset

**Table 10. Comparison with other character recognition methods on TUNISIAN dataset**

| Recognition Performance (%) | | |
|---|---|---|
| Subset<br>Method | TLPs1 | TLPs2 |
| **Ktata *et al.* [54]** | 90.7 | - |
| **DELP-DAR (ours)** | 96.9 | 97.5 |

of few characters, due to the high completeness of images or of total failure of letters in LPs (see Fig 13). Another problem noticed in some images is the confusion of characters, for example between letter O and digit 0, or between letter I and digit 1.

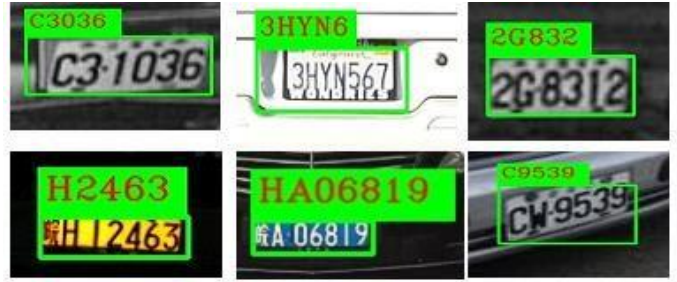

**Fig. 14.** Example of failure characters recognition

## 5. Conclusion and future work

Several research studies have addressed LP detection and recognition. In fact, many authors have implemented several approaches and techniques for the application of this process.

However, all the methods have their own advantages and disadvantages. Moreover, each country has its own system of numbering LPs: backgrounds, sizes, colors and languages of characters.

This research is different from previous work: On the one hand, the use of object detection techniques and approaches like Mask-Rcnn and their adaptation of this approach for the detection of specific objects, as well as for the detection and recognition of characters, is not an easy task because of the different size and multilanguage text.

On the other hand, the new creation of the Tunisian dataset is a major contribution in our work so that researchers can test their codes and evaluate their results with a new dataset, especially that containing Arabic letters.

Our DELP-DAR framework is based on three parts: The first one is to detect LPs using the Mask-RCNN with some modifications.

The second part is to segment LPs, mainly detecting characters. The third one is to recognize both latter correctly.

Our proposed method has been extensively evaluated on the widely used Caltech, AOLP, PKU datasets as well as our new Tunisian dataset.

The experiments have demonstrated that our proposed approach substantially outperforms the state-of-the-art methods in terms of recognition accuracy in all datasets, but in LP detection there is some failure compared to a single author in the PKU and AOLP datasets only.

As demonstrated in the experiments, the proposed approach and the other state-of-the-art methods are still subject to certain limitations when addressing difficult scenes, low resolution, terrible illumination and accidental occlusion. In the future, it will be interesting to add some techniques to address such difficult cases.

Another limitation in our work is the speed of our framework, which we do not consider in our framework due to the weakness of the performance of the used equipment. For this, and in future work, we will try to add the criterion of speed in our experimentation.

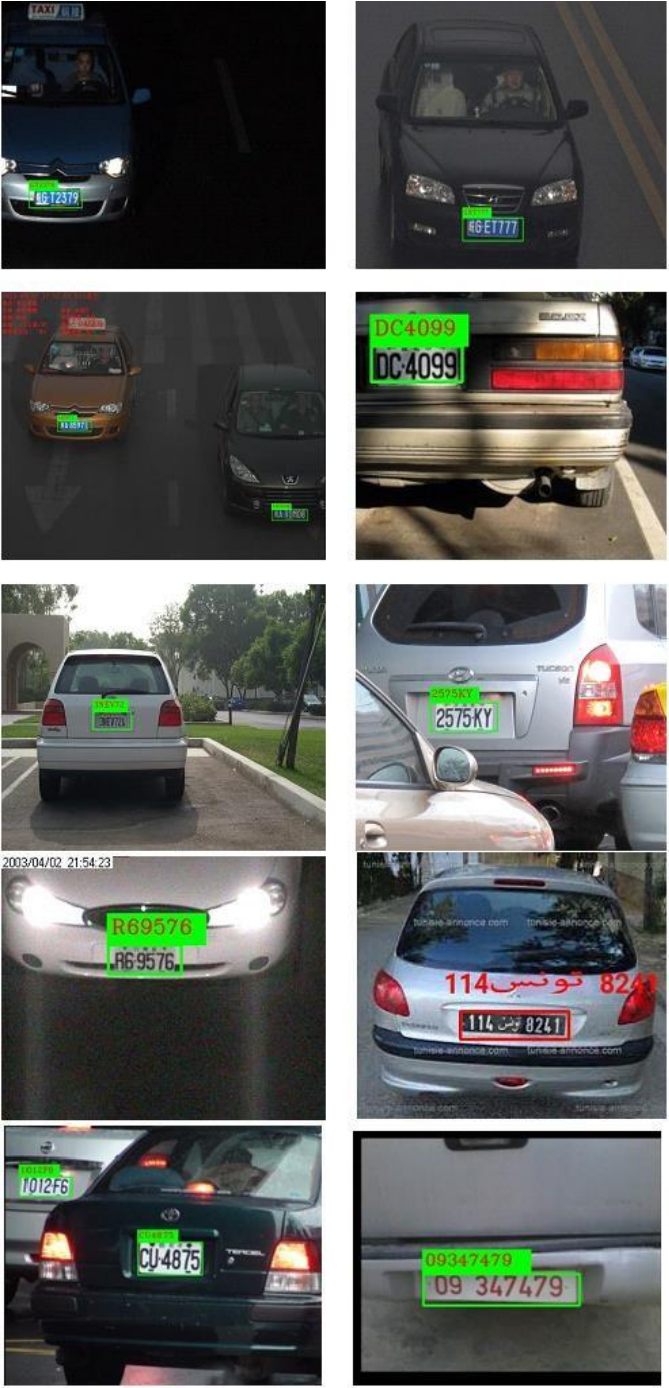

**Fig. 12.** Example of LP recognition (the red boundary boxes is the new Tunisian LP)

*4.6. Analysis and Discussion for character recognition*

Our approach of LP recognition is robust and efficient compared to other studies in the literature. Consequently, our obtained results are better. This robustness is due to the good segmentation method, which gives us very high values and a very low error rate.

The correct classification of the characters in Caltech, PKU, AOLP and Tunisian datasets, as shown in Fig.12, is done with the robustness of the used model, as well as the number of characters used in the training phase. The increase in the number of characters gives us an improvement in the result. It is the used data augmentation technique that helps us to achieve such a result. Furthermore, in this phase we notice the failure of the recognition

In the same vein, we will try to develop our real-time System using new technologies (smartphones, tablets, etc.) so that we can exploit the DLLP-DAR framework in a mobile environment.

## 6. ACKNOWLEDGMENT

The research leading to these results has received funding from the Ministry of Higher Education and Scientific Research of Tunisia under the grant agreement number LR11ES4.